\documentclass[conference]{IEEEtran}
\usepackage{cite}
\usepackage{amsmath,amssymb,amsfonts}
\usepackage{algorithmic}
\usepackage{graphicx}
\usepackage{textcomp}
\usepackage{xcolor}

\usepackage{hyperref}
\hypersetup{hidelinks}

\def\BibTeX{{\rm B\kern-.05em{\sc i\kern-.025em b}\kern-.08em
    T\kern-.1667em\lower.7ex\hbox{E}\kern-.125emX}}

\begin{document}

\title{Predicting dominant hand from spatiotemporal context varying physiological data}

\author{\IEEEauthorblockN{Jorge Neira-García, and Sudip Vhaduri}
\IEEEauthorblockA{\textit{Polytechnic Institute}\\
\textit{Purdue University}\\
West Lafayette, Indiana, USA \\
Emails: jneiraga@purdue.edu, svhaduri@purdue.edu}
}

\maketitle

\begin{abstract}
Health metrics from wrist-worn devices demand an automatic dominant hand prediction to keep an accurate operation. 
The prediction would improve reliability, enhance the consumer experience, and encourage further development of healthcare applications. 
This paper aims to evaluate the use of physiological and spatiotemporal context information from a two-hand experiment to predict the wrist placement of a commercial smartwatch.
The main contribution is a methodology to obtain an effective model and features from low sample rate physiological sensors and a self-reported context survey. 
Results show an effective dominant hand prediction using data from a single subject under real-life conditions.
\end{abstract}


\section{Related Work / Literature Review} 
Smartwatches are getting higher technology for increasingly affordable prices and their popular health-tracking applications demand accuracy \cite{Reeder2016}. 
Alternatives to overcome wrist placement inaccuracies are required to support well-informed health promotion for smartwatch consumers \cite{Buchan2020}. 
Automatic dominant hand prediction would reduce negative impacts from an incorrect setup process or wrist placement changes. 
The spatiotemporal user context and its connection with physiological variations provide a potential solution for dominant hand prediction and accuracy improvements.

Addressing concerns in terms of health impact, three popular commercial wrist-worn devices were recently reviewed in \cite{Hajj2022}.
Results showed accurate heart rate measurements, but poor energy expenditure estimations.
It is consistent with results for previous models presented in \cite{Wallen2016}, six years ago.
Professionals and users are warned to proceed with caution using those devices for training or nutritional purposes.
Manufacturers are asked to improve algorithms addressing potential error sources.

Device placement is one of the factors influencing the accuracy of accelerometer data during human activity recognition \cite{Davoudi2021}.
In particular, wrist-worn devices are known to be susceptible to dominant or non-dominant wrist placement with up to 8.5\% variations \cite{Buchan2020}.
Some manufacturers seem aware of this problem and claim to include measures that counteract its effects, see \cite{fitbit_myhelp}.
However, the corrective measures depend on user input during initial setup or setting adjustments \cite{fitbit_2022}.
An automatic dominant hand prediction would improve the user experience, with added reliability towards accurate health information.

Developing an effective automatic solution faces various challenges to detect a dominant hand.
Only low sample rate sensor information is available from consumer-grade devices.
That limits elaborated physiological measures such as Heart Rate Variability (HRV) requiring up to 250 Hz \cite{Kwon2018}.
The prediction should also work under real-life unconstrained conditions.
However, the dominant hand is only distinguished in activities including writing, eating, cooking, dishwashing, or generally using tools demanding strength and precision \cite{NICHOLLS20132914}.

Machine learning algorithms are already used for different activity recognition tasks \cite{Weiss2016,Vaizman2017}.
So, merging human context and sensor data from different sources is a strategy with proven potential to deal with the established challenges.
A remarkable example effectively estimates stress levels using limited physiological data and activity information to improve its performance \cite{Can2020}.

\section{Approach / Method}
The objective is to use machine learning concepts to predict the dominant hand of a user wearing a commercial smartwatch.
A two-hand experiment was proposed to gather physiological data simultaneously from both hands for over 2-weeks.
A survey was designed for the user to keep a self-report of contextual information including location, activity, and subjective physiological arousal.
The raw data was downloaded using the official export tools from the manufacturer.
Algorithms using python and Matlab were proposed to clean, analyze and calculate features from the raw data.
Finally, using feature selection algorithms a relevant subset was used to train and select an effective classification model.

\subsection{Two-hand experiment conditions}
Smartwatches are getting higher technology sensors and processing units for increasingly affordable prices.
Sensors commonly include GPS, gyroscopes, pedometers, temperature, blood oxygen (SpO2), and advanced heart monitoring. 
The available data enables gesture recognition, physical activity identification, and estimation of relevant health metrics. 
These wearable devices are now a popular alternative to easily monitor wellness information, such as energy expenditure, sleep quality, heart rhythm assessment, and stress levels.  

\begin{figure}[t]
\centerline{\includegraphics[width=\columnwidth]{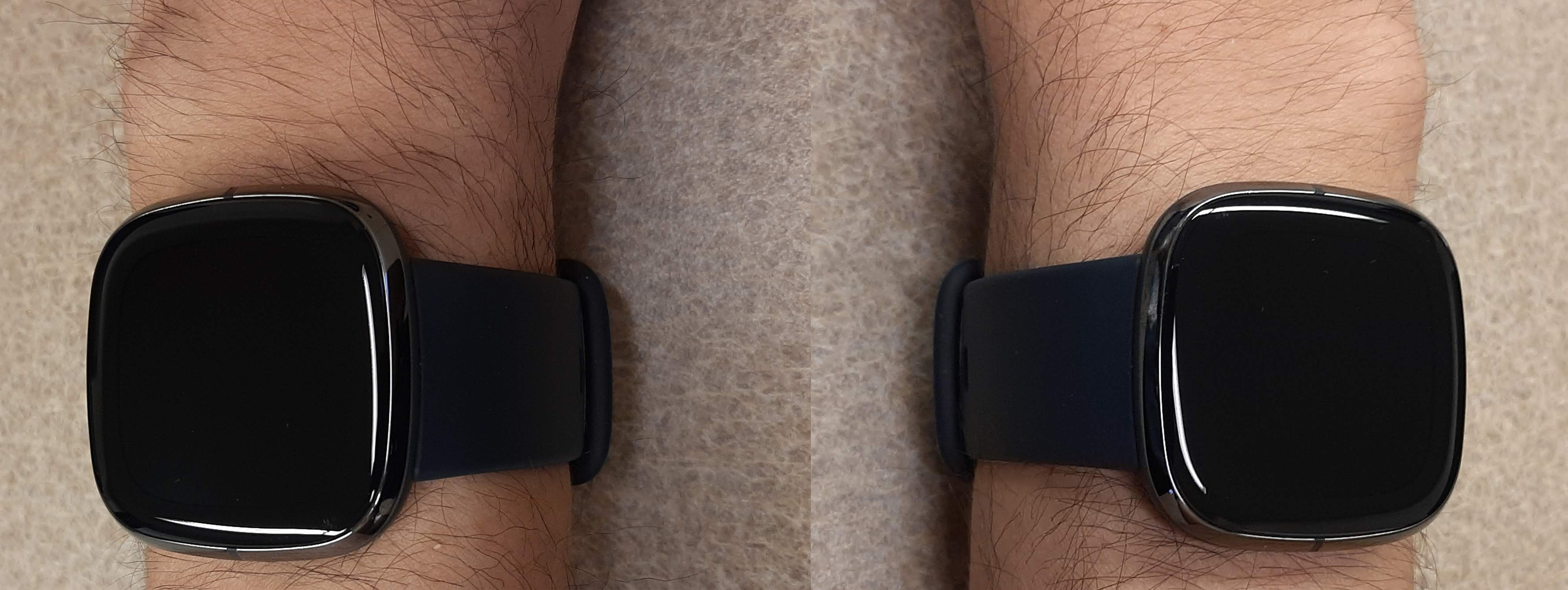}}
\caption{Smartwatch placement for two-hand experiment}
\label{fig:FitbitPhoto2Hand}
\end{figure}

The proposed two-hand experiment takes advantage of those characteristics to develop the dominant hand prediction algorithm.
The subject wore a Fitbit Sense device on each wrist following its recommended tightness and location for all-day wear with one finger width above the wrist bone, see \autoref{fig:FitbitPhoto2Hand}.
Given the long 2-week time frame, the smartwatches were removed to keep consistent measurement conditions. 
This was done for recharging the devices, a daily cleaning procedure, and also to prevent interaction with water.

After the initial setup, the settings were kept with default values.
All the available features were turned on, and the devices had an active Fitbit Premium subscription during the experiment.
Each wristband was connected to a different smartphone due to a one-device restriction for the official application.
Both device pairs were kept reasonably close to ensure a consistent Bluetooth link during the experiment.

Lastly, it is important to highlight that the subject was instructed to keep routines unchanged.
Other than wearing the pair of smartwatches, keep them charged, and clean.
The data should record typical unconstrained real-life conditions for a graduate student without mobility restrictions.

\subsection{Self-report survey for contextual information}
\begin{figure}[b]
\centerline{\includegraphics[width=0.8\columnwidth]{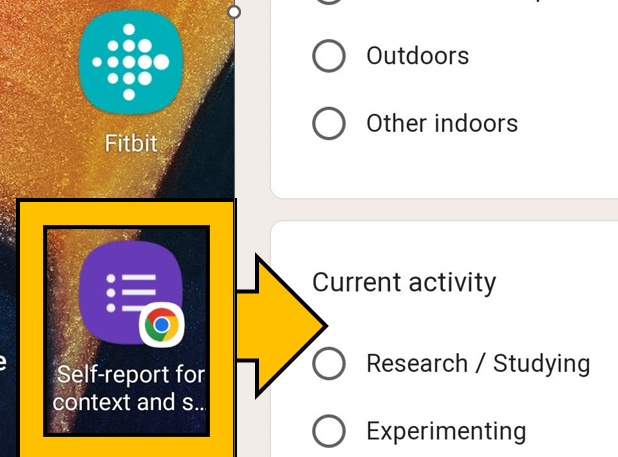}}
\caption{Self-report survey implementation}
\label{fig:FitbitSelfReport}
\end{figure}

A simple and short survey was designed to easily gather detailed information on the subject's context. 
The user was prompted to make only between 1 to 4 selections among different options of interest. 
Those options included 10 common locations, 16 activities, 5 levels of physiological arousal, and a flag to indicate when the device was removed. 
The survey was implemented using Google Forms as shown in \autoref{fig:FitbitSelfReport}.

The suggested procedure was filling out the survey before any change of activity or location. 
A shortcut access was added on the home screen of the subject's main smartphone to remember the self-report task. 
However, the subject remarked serious difficulties to keep on the self-report during the experiment. 
It was difficult to establish the routine, the task was easily forgotten even with an expressed high level of engagement with the experiment. 
Unreliable internet connection and slow smartphone responses further contributed to the problems. 

The survey demanded unrealistic commitment under a highly dynamic real-life routine. 
So, the subject completed some information using the survey, but also manually included additional entries to complete the spreadsheet at the end of each day.
The manual updates considered the inspection of additional sources of information to increase accuracy on the context labels.
That includes using GPS information from Google Timeline records, heart rate, and step count from the Fitbit dataset.
During the experiment 256 context updates were completed using the survey, and 241 additional updates were included manually. 

It is worth clarifying that the purpose of the survey is just to facilitate the research process.
However, a fully practical implementation would require the extraction of contextual information from the available sources.
For instance, location can be automatically extracted from GPS information, and activity can be estimated from the numerous sensors on both smartphone and smartwatch devices.

\subsection{Data processing}
The process starts using the official data export tool from the Fitbit website to download the most detailed raw data available.
Files for heart rate, step count, calories, and altitude are stored in separate folders for each hand.
The manufacturer encodes the information in multiple JSON-formatted files.

A python code with inspiration on \cite{ottesen_2019} is developed to convert the JSON files into a single CSV file for each hand.
The CSV files are further processed using Matlab to remove unnecessary information, adjust nonuniform date-stamps, and re-sampling to a consistent 1-second interval dataset.
Missing information was filled using linear interpolation for heart rate, and zero value for the remaining variables.
Clean and consistent datasets for each hand are stored as timetables on MAT files.

For the self-reported contextual information.
The spreadsheet containing the data is directly processed using Matlab.
A timetable structure is also created and string data types are converted into categorical.
A MAT file is saved with a re-sampling of a 1-second interval consistent with the physiological data.
Another one is saved with a greater re-sampling under the window interval used for feature calculation. 
A new time categorical column is created to divide time into four more meaningful ranges (Noon, Morning, Afternoon, and Evening).

Time synchronization is then performed using the Matlab timetables to merge both physiological and contextual data with the uniform 1-second intervals. 
The physiological data is further cleaned eliminating time frames where the Fitbit devices are removed. 
Finally, the corresponding dominant and non-dominant hand labels are assigned to the clean, uniform, and organized datasets ready for feature extraction.

\subsection{Feature Extraction}
\begin{figure}[b]
\centerline{\includegraphics[width=\columnwidth]{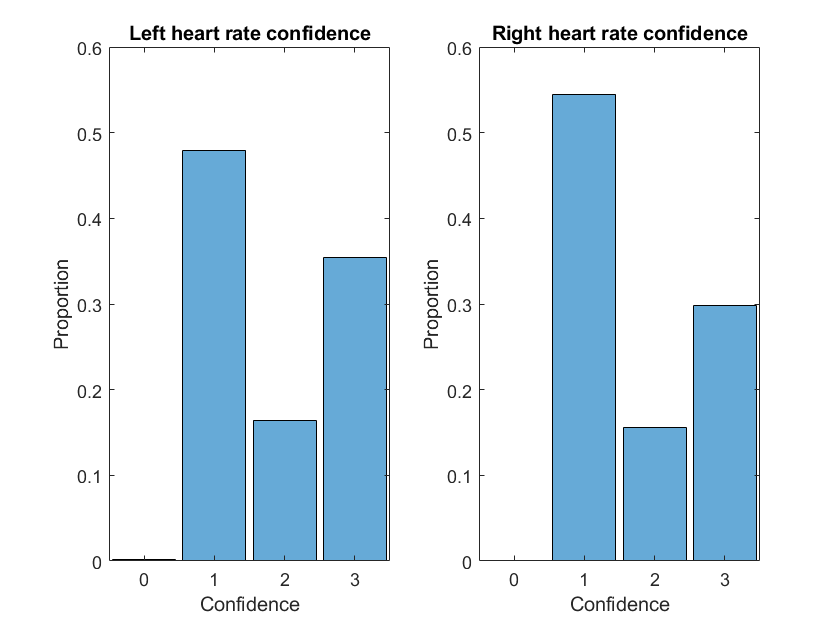}}
\caption{Histogram for heart rate confidence values}
\label{fig:HR_Confidence_Histogram}
\end{figure}

The objective on dominant hand prediction was clearly defined only after the experiment design and its data collection.
So, some of the collected data is dropped as irrelevant given the context of this problem.
For instance, altitude from the Fitbit data, and subjective physiological arousal form the self-report survey.
The confidence for each heart rate sample was also omitted given the similar distribution of its values for both dominant and non-dominant hand, see \autoref{fig:HR_Confidence_Histogram}.

The features for the remaining data were calculated using different window sizes of 1, 5, 10, 20, and 40 minutes.
For heart rate 3 frequency-based, and 17 time-based statistical features are calculated.
The cumulative sum was the only feature considered for step count and calories.
The 3 categorical features were not further processed and completed a set with 25 features in total.

The experiment was extended two additional weeks to test two different settings available on Fitbit devices.
The first one with both devices on default non-dominant hand setting.
The second one with the devices configured accordingly to the dominant or non-dominant hand setting.
The features are calculated for both conditions and stored in separated files.

\subsection{Context Information and Feature Selection}
The Minimum Redundancy Maximum Relevance (MRMR) Algorithm, available on Matlab, is initially used for feature selection.
Following an heuristic approach, it provides significant scores for classification tasks with both categorical and continuous features.
The results show low relevance for most of the heart rate and categorical features.
\begin{figure}[t]
\centerline{\includegraphics[width=\columnwidth]{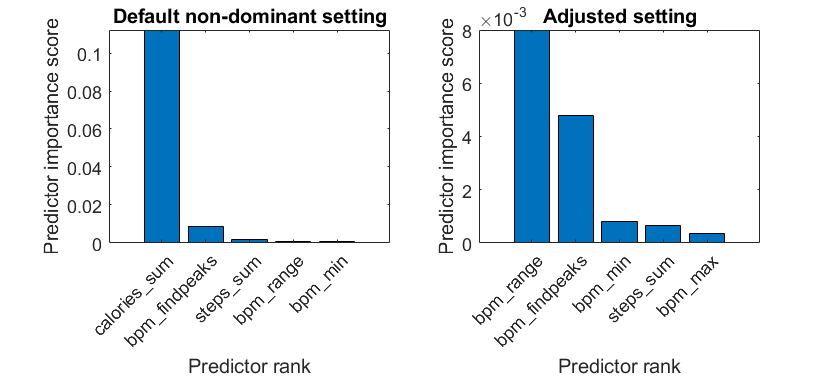}}
\caption{10-minute window features ranked by MRMR score}
\label{fig:MRMR_Features10m}
\end{figure}

For the example shown in \autoref{fig:MRMR_Features10m}, features with top-five MRMR scores include calories, step count, and heart rate values for range, minimum, maximum and number of peaks.
The results vary depending on the window size, but also on the Fitbit setting being used, something evidenced on the example shown.
However, it is important to remark that most categorical features got low or zero scores on almost every tested condition.

Notwithstanding the MRMR scores, categorical features contain some valuable contextual information such as activity, location, and time.
From literature, see \cite{NICHOLLS20132914}, it is clear that certain activities are strongly connected with dominant hand preference.
Therefore, a further feature filter accounting to context and problem knowledge is proposed.

The context-based feature filter eliminates windows potentially providing indistinguishable data for dominant prediction.
That includes activities involving reduced body movement, such as sleeping, movies, or meetings.
Physiological data suggesting similar conditions, like zero step or zero calories counts.
Also, including means to filter windows with specific activities or conditions facilitating dominant hand prediction.

\subsection{Classification algorithms}
The Matlab Classification Learner toolbox was used to evaluate a broad variety of models.
It includes up to 32 different model configurations such as decision trees, discriminant analysis, logistic regression, naive Bayes, support vector machines (SVM), k-nearest neighbor (KNN), kernel approximations, ensemble alternatives, and neural networks.
The models supporting both categorical and continuous features are strongly limited, but most of the tests included continuous features only.
The main configurations across all the different evaluations were set to a 5-fold cross-validation method and 10\% proportion for the test dataset.
Access to the data and codes is available on the following link: \url{https://drive.google.com/drive/folders/1kdaSetBQWjWsV2pbKZ-Y41dZirmyWskS?usp=sharing}

\section{Dataset Description}
Two sets with a commercial smartphone paired with a Fitbit smartwatch were used for the project.
All the information is gathered for a single individual without constraints on its daily life behavior.
The devices record timestamped physiological data with a low second sample rate and the following specifications for each hand:
\begin{itemize}
    \item Heart Rate (HR) ($\sim$355000 data points with 7 s average sample time)
    \item Steps ($\sim$19000 data points with 2 min average sample time, only 5000 non-zero entries)
    \item Calories ($\sim$43000 data points with 1 min average sample time)
    \item Altitude ($\sim$140 data points with 4 h average sample time)
\end{itemize}
The self-report survey records include 497 data points for:
\begin{itemize}
    \item Wearing/Removing smartwatch flags
    \item Location (10 common points of interest)
    \item Activity (16 common options)
    \item Subjective physiological arousal (5 levels)
\end{itemize}


\section{Analysis/Results/Evaluations}
The baseline evaluation considered default non-dominant Fitbit setting for both hands, a window size of 1 min, 4 out of 26 top-scoring features using MRMR, all the recorded data without context-based feature filter.
From the 32 classification models, neural networks, SVM, and Naive Bayes required the highest training time reaching over the 500 s mark.
The top-5 validation accuracy models are decision trees and ensemble versions reaching up to 87.8\% accuracy.

The results suggest that some feature is providing biased information.
From the unconstrained real life context, it is not expected to easily identify the dominant hand under most of the activities.
From the MRMR scores the calories count have a value around 100 times higher compared to the next feature in the list, the peak counts for heart rate.
A test omitting the calories count for the highest ranked classifier further confirms the impact of the feature on model performance dropping best accuracy to 50.4\%.
The following evaluations omit the calories feature, and change other conditions to reveal a more realistic result.
\begin{figure}[b]
\centerline{\includegraphics[width=0.7\columnwidth]{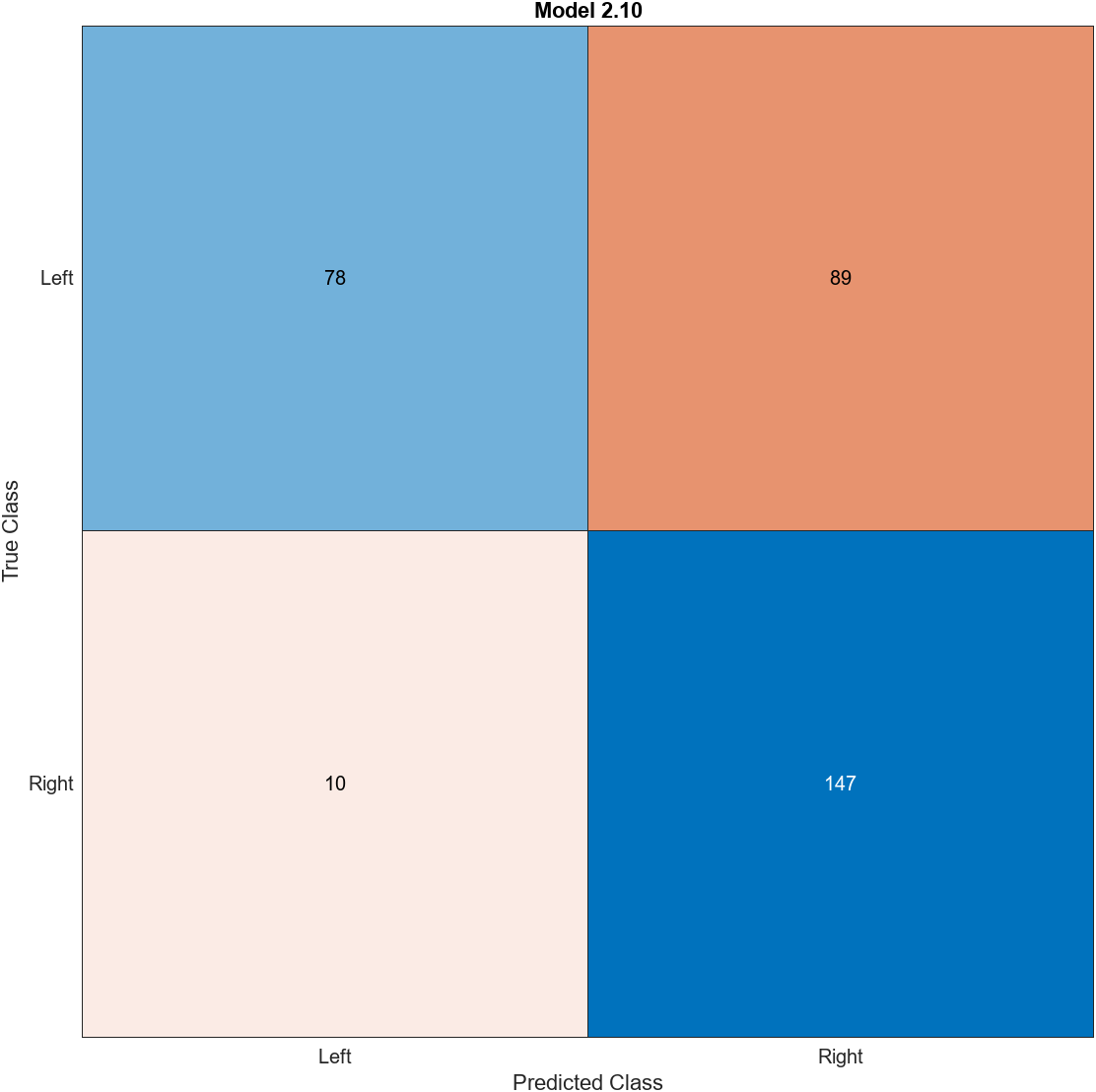}}
\caption{Confusion Matrix for Quadratic SVM 1 min window, exercising only}
\label{fig:QuadraticSVM_Exc_1m_ValidationCM}
\end{figure}

Another study evaluates the value of using the proposed context-based filter to clean the dataset.
It is performed only using data from all activities with potential to identify dominant hand, and with step-counts different from zero.
The results show a consistent accuracy between validation and test data for the Coarse Decision Tree and the Ensemble Boosted Decision Tree reaching up to 59.5\%.
The improvement on classification performance, motivated additional individual evaluations for different relevant activities.

The study for individual evaluations start with the exercising activity.
It is worth mentioning that MRMR feature selection is used and 3 out of 21 top-scoring features are chosen.
The top features usually include heart rate peak count, step-count, and an statistical heart rate measure like range.
For these conditions, a quadratic SVM results with consistent accuracy around 70\% for both validation (see \autoref{fig:QuadraticSVM_Exc_1m_ValidationCM}) and test data.
For other individual activities such as doing chores, and walking lower accuracy of 58.6\% and under 50\% values are respectively obtained for the best performing models.

\begin{figure}[t]
\centerline{\includegraphics[width=0.7\columnwidth]{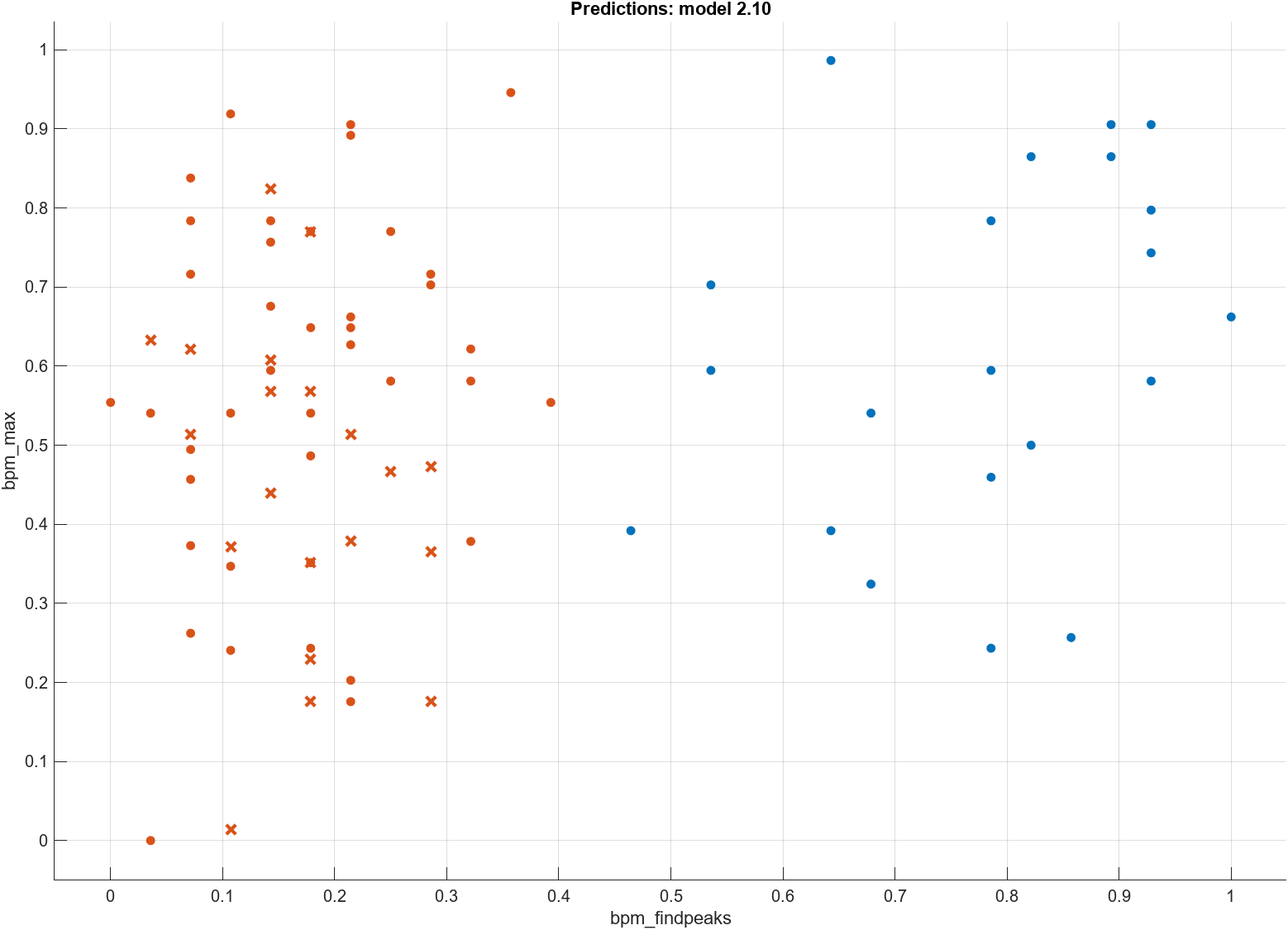}}
\caption{Scatter plot for features used with 5 min window evaluation}
\label{fig:ScatterPlot_5mWindow}
\end{figure}

The remaining evaluations are performed as modifications from the best performing conditions.
Using the context-based filter to work with only exercising activity, without zero step-counts, ignoring calories count, and selecting top-scoring features with MRMR.
Evaluating the effect of using a wider window size. 
The window was changed from 1 min to 5 min.
As a result, MRMR selection showed a dominant score for the heart rate peak counts.
Training models with just two features resulted in accuracy values near 80\% for quadratic SVM, but scatter plots evidence almost exclusive contribution from the peaks feature, see \autoref{fig:ScatterPlot_5mWindow}.


\section{Discussion / Limitations}
From the results it is clear that a dominant hand prediction can be obtained with up to 70\% accuracy.
It is critical to carefully select the available physiological and contextual information.
In particular, the results were better obtained when the individual is exercising.
The MRMR algorithm recognize 3 relevant features including step-count, heart rate peak count, and range.

It is important to highlight that only data for one person was used across the study.
The data was collected under no activity restrictions, and detailed information for the most relevant activities on dominant-hand prediction was strongly limited.
The labels for the evaluated activities were manually recorded and not detailed.
The physiological information has low sample rates and it is limited to manufacturer logs.

\section{Conclusion \& Future Work}
The impact and importance of contextual information is evidenced through the different analysis.
Dominant hand detection seems viable using limited commercial smartwatch data and it is constrained to reliable activity detection.
There is a long list of further tests and aspects to consider, some including:
\begin{itemize}
    \item Further model selection, speed concerns
    \item Improved data processing and feature calculation
    \item Sensor fusion Smartphone Step-count
    \item Significant activity selection
    \item Automatic activity recognition
    \item Record and usage of accelerometer data
    \item Experiments collecting data for more persons
    \item Additional impact and mitigation of inaccuracies when dominant hand is detected
\end{itemize}



\bibliographystyle{IEEEtranTIE}
\bibliography{conference_101719}\ 

\end{document}